\documentclass[11pt]{article}

\usepackage[final]{acl}

\usepackage{times}
\usepackage{latexsym}

\usepackage[T1]{fontenc}

\usepackage[utf8]{inputenc}

\usepackage{microtype}

\usepackage{inconsolata}

\usepackage{graphicx}

\newcommand{\tp}{t^+\xspace}

\usepackage{xspace}
\usepackage{amsmath}
\usepackage{amsfonts}
\usepackage{multirow}
\usepackage{booktabs}
\usepackage{tabularx}
\usepackage{wrapfig}
\usepackage{subcaption}
\usepackage{hyperref}
\usepackage{wrapfig}
\usepackage{xcolor}
\usepackage{fancyvrb}
\usepackage{multirow}
\usepackage{enumitem}
\usepackage{graphicx}

\title{A Temporal Knowledge Graph for Music Festival Lineup Forecasting}

\author{Julia Gastinger, Thilo Dieing, Christian Meilicke, Heiner Stuckenschmidt  \\
  University of Mannheim,   Mannheim, Germany \\
  \texttt{first.last@uni-mannheim.de} \\ 
  }

\begin{document}
\maketitle

\begin{abstract}
Music festival lineups emerge from complex relationships among artists, genres, releases, labels, and past performances, making the prediction of future lineups a natural fit for temporal knowledge graph (TKG) forecasting.
In this work, we present a TKG covering 380 festivals over 55 years, comprising more than 90K festival performance quadruples along with information on festivals, artist tours, and artist metadata, and release it as a resource for TKG forecasting evaluation.
We formalize festival lineup forecasting as temporal link prediction between artists and festivals at future timestamps.
We evaluate six TKG forecasting models on this task, analyze their capabilities and limitations, and compare them against Large Language Models applied zero-shot.
Our resource complements existing TKG benchmarks by grounding evaluation in a concrete, real-world application domain.
\end{abstract}

\section{Introduction}
Temporal knowledge graphs (TKG) incorporate temporal information into knowledge graphs (KG)~\cite{Han2021xerte}. The task of TKG forecasting, which aims to predict future links from historical information, has recently gained increasing attention and motivated the development of a variety of approaches~\cite{Li2021regcn, Liu2021tlogic,chen2025cogntke}.

Alongside these modeling advances, recent work has begun to address the evaluation of TKG forecasting, proposing a unified evaluation protocol, a heuristic baseline, large-scale datasets, and an evaluation framework~\cite{gastinger2023eval,gastinger2024baselines,gastinger2024tgb}. 

Yet evaluation has so far relied on benchmarks not tied to any concrete application scenario, and the resulting prediction tasks do not necessarily correspond to questions a practitioner would meaningfully want to answer.  
Most established benchmarks are extracted from political event data and news sources via automated NLP pipelines (e.g., ICEWS~\cite{Boschee2015} and GDELT~\cite{Leetaru2013GDELT}) or based on general-purpose knowledge bases (e.g., Wikidata~\cite{vrandevcic2014wikidata}, YAGO~\cite{Mahdisoltani2015YAGO}). 
Creating TKGs from event extraction pipelines can result in queries that cannot be interpreted without additional contextual knowledge. The TKGs derived from knowledge bases often include facts that remain stable over long period, which results in a task that primarily requires repeating previously stated facts rather than forecasting~\cite{gastinger2024baselines}.

These characteristics reflect the gap between general-purpose evaluation and the needs of application scenarios. 
This motivates the development of complementary resources grounded in concrete use cases, with well-defined entities, interpretable relations, and realistic temporal dynamics.

Music festival lineups are shaped by complex interactions among artists, festivals, genres, labels, and past performances. \citet{hiller2016quality} shows that hiring decisions depend on factors such as an artist's past festival appearances, album quality, and touring history across multiple years. 
This makes music festival lineups and the related interactions a natural fit for temporal relational modeling.
Beyond their suitability as a research domain, music festivals represent a significant cultural and economic sector. In Germany alone, 
the festival landscape generates estimated annual revenues of €551 Mio., 
with artist fees accounting for approximately 38\% of expenditures~\cite{festivalstudie2025}. 
Consequently, predicting future festival lineups is directly relevant to different groups within the festival ecosystem: It can support organizers in curating lineups, help artists and managers identify suitable events, and assist fans in planning attendance decisions, especially since many festivals sell early-bird tickets before full lineups are revealed.

In this work, we construct a TKG of music festival lineups as benchmark for TKG forecasting models. We focus on the following contributions:
\begin{itemize}[leftmargin=*]
\setlength{\itemsep}{0pt}
\setlength{\parskip}{0pt}
\setlength{\parsep}{0pt}
    \item We provide a TKG spanning 55 years and 380 festivals, integrating festival lineups, concerts, venues, and artist metadata, along with a documented, reproducible construction pipeline.
    \item We introduce the task of lineup forecasting. It focuses on relations of interest for the given use case, and covers two query directions (artist-prediction and festival-prediction). In this context, we argue for two metrics better suited to the many-correct-answer structure of lineup queries.
    \item We conduct an empirical evaluation of six forecasting models and zero-shot LLMs, analyze behavioral differences, and discuss strengths and open issues. 
\end{itemize}

\vspace{-1em}
\paragraph*{Resource Availability:} 
    Source code, including dataset creation, evaluation,  tested baseline implementations, a lightweight ontology file, an RDF-star version of our dataset, and Croissant metadata are available on \url{https://github.com/JuliaGast/festival-lineup-tkg}.
    The final TKG datasets are available on \url{https://madata.bib.uni-mannheim.de/822/}. 

\section{Prediction Task}\label{sec:task}
The task in this work is to predict future lineups of music festivals based on historical information.
We model this historical information as a temporal knowledge graph, and the task of predicting future festival lineups as a TKG forecasting task, i.e., the task of predicting future links in a TKG.

A temporal knowledge graph~$G$ is a set of quadruples $(s,p,o,t)$ with entities (or nodes)
$C(G) = \{ s \mid (s,p,o,t) \in G \} \cup \{ o \mid (s,p,o,t) \in G \}$, relations $P(G) = \{ p \mid (s,p,o,t) \in G \}$, and timestamps $T(G) = \{ t  \mid (s,p,o,t) \in G \} \subset \mathbb{N}$. Timestamps may represent hours, days, years, or any other temporal granularity, depending on dataset and use case. In our case, timestamps are years.
The semantic meaning of a quadruple (also referenced as a temporal triple) $(s,p,o,t)$ is that $s$ is in relation $p$ to $o$ at time $t$. An example is \textit{(Arlo Parks, performs at festival, MS Dockville, 2023)}.

Given a TKG $G$, TKG forecasting or extrapolation is the task of predicting quadruples $(s,p,o,t^\star)$ for future timestamps $t^\star > max(T(G))$ with $p \in P(G)$ and $s,o \in C(G)$. In our project, we will focus on the task of {entity forecasting}, that is, predicting object or subject entities for queries $(s,p,?,t^\star)$ or $(?,p,o,t^\star)$. Akin to static knowledge graph completion, TKG forecasting is approached as a ranking task \cite{han2022TKG}. For a given query, which is usually derived from a quadruple in a test or validation set, a model needs to rank entities in $C(G)$ using a scoring function that assigns plausibility scores~\cite{gastinger2025ruc}. 

In our work, we ask two specific types of queries: \textit{(A, performs at festival, ?, $t^\star$)}, i.e. at which festival will a given artist $A$ play at in a future timestamp $t^\star$, and \textit{(?, performs at festival, F, $t^\star$)}, i.e. which artist will perform at a given festival $F$ at a future timestamp $t^\star$. We also refer to these two types of queries as festival prediction and artist prediction.

We focus on single-step prediction, which means we predict one timestamp into the future. Since our dataset has yearly granularity, this means we predict the festival lineup for the next year.

\section{Related Work}
\label{sec:rel_work}
\subsection{TKG Forecasting Models}
A wide range of methods has been proposed for temporal knowledge graph forecasting. One prominent line of work combines graph-based message passing with sequential modeling to capture structural and temporal patterns; representative models include RE-Net~\cite{Jin2020renet}, RE-GCN~\cite{Li2021regcn}, xERTE~\cite{Han2021xerte}, TANGO~\cite{Han2021tango} 
and CEN~\cite{Li2022cen}. Further approaches in this direction include history-based prediction (CyGNet~\cite{Zhu2021cygnet}), contrastive learning (CENET~\cite{Xu2023CENET}), and combined local and global graph encoding (TiRGN~\cite{Li2022TiRGN}).
A second direction uses reinforcement learning to discover temporal reasoning paths, with 
TimeTraveler~\cite{Sun21TimeTraveler} as representative example.

Rule-based methods aim to learn interpretable temporal logic rules. TLogic~\cite{Liu2021tlogic} learns such rules via temporal random walks, while CountTRuCoLa~\cite{gastinger2025ruc} learns temporal confidence functions that account for frequency and recency. 
Hybrid approaches such as INFER~\cite{li2025iinfer} and CognTKE~\cite{chen2025cogntke} combine rule-based and embedding-based methods. More recently, LLM-based models have been proposed to incorporate semantic knowledge into forecasting, including zrLLM~\cite{ding2024zrllm} 
and GenTKG~\cite{liao2024gentkg}.
Finally, the Recurrency Baseline~\cite{gastinger2024baselines} is a heuristic baseline that predicts fact recurrence by incorporating temporal distance and frequency.

\subsection{TKG Forecasting Benchmark Datasets}
Most established benchmark datasets are derived from political and news event data, including the Integrated Crisis Early Warning System (ICEWS)~\cite{Boschee2015,shilliday2012data}, the Global Database of Events, Language, and Tone ({GDELT})~\cite{Leetaru2013GDELT}, and POLECAT~\cite{Scar2023polecat}. 
In addition, some datasets are constructed from knowledge bases such as Wikidata~\cite{vrandevcic2014wikidata} and YAGO~\cite{Mahdisoltani2015YAGO}. 

\textbf{Limitations of Benchmarks:}
While recent work has improved the evaluation of TKG forecasting methods~\cite{gastinger2023eval,gastinger2024baselines,gastinger2024tgb}, the resulting tasks can remain difficult to interpret from an application perspective: evaluation is typically framed as generic future link prediction  aggregated over all relations~\cite{gastinger2024tgb}, often spanning a wide variety of domains within a single dataset.
Several issues compound this:
\begin{enumerate}[
    leftmargin=1.05em,
    labelsep=0.4em,
    itemsep=0pt,
    topsep=0pt
]
\setlength{\itemsep}{0pt}
\setlength{\parskip}{0pt}
\setlength{\parsep}{0pt}
    \item Test queries are sometimes unintuitive without access to the full contextual knowledge. For example, \texttt{tkgl-polecat} contains triples such as \textit{(Ben Cardin, THREATEN, Federation)} that are impossible to interpret in isolation. 
    \item Automatic extraction from news text produces relations with heterogeneous domains and ranges. For example, in ICEWS-based datasets, the relation \textit{Arrest, detain, or charge with legal action} connects countries to demographic groups (e.g., 
    \textit{Children (France)}), abstract roles (e.g., \textit{Citizen (Australia)}, \textit{Criminal (Philippines)}), and individual persons (e.g., \textit{Andal Ampatuan Sr}), complicating interpretation and model analysis.
    \item Wikidata-derived benchmarks such as \texttt{tkgl-smallpedia} and \texttt{WIKI} include many queries on facts that are stable across decades (e.g., religion, administrative capitals). More than 
    $70\%$ of test facts in these benchmarks were already true at their previous timestep and a simple recurrency-based baseline is competitive with or superior to most TKG forecasting methods on them~\citep{gastinger2024baselines,gastinger2024tgb}. This suggests that high stability leaves limited room above simple persistence.
    \item Some datasets contain relations that cannot be reliably inferred from available history. 
    \texttt{YAGO}, for instance, provides insufficient context to forecast prize awards or deaths~\cite{gastinger2024baselines}. 
    \item The news-based \texttt{GDELT} contains not only real-world dynamics, but also media reporting cycles: Its 15-minute resolution leads to repeated mentions of the same event across adjacent timestamps~\cite{gastinger2025ruc}, making it difficult to determine what models actually learn.
\end{enumerate}
These points suggest that benchmark scores may not always reflect a model’s ability to answer practically relevant forecasting queries. 
This motivates the development of resources grounded in concrete use cases with well-defined entities, understandable relations, and realistic temporal dynamics.

\section{TKG Creation}
\subsection{Data Sources}
We extract information from two sources: \texttt{setlist.fm}~\cite{setlist26} and \texttt{MusicBrainz}~\cite{musicbrainz26}.
\textbf{\href{https://www.setlist.fm/}{\texttt{setlist.fm}}} is a community-maintained wiki-style platform for collecting and sharing concert setlists. Beyond individual concert setlists, it contains information on venues, festivals, and the artists who play there. The platform provides a public API for data extraction, with all artists, festivals, concerts, and venues assigned unique identifiers. In \texttt{setlist.fm}, artists are represented by MusicBrainz identifiers. This enables straightforward cross-referencing between the two data sources.
\textbf{\href{https://musicbrainz.org/}{\texttt{MusicBrainz}}} is an open music encyclopedia that collects and publishes music metadata, maintained by a global community of contributors. Like \texttt{setlist.fm}, it offers an API for straightforward data retrieval.

\subsection{Data Mining and Reduction}
We initially fetch information on all festivals, their editions, venues, countries, and associated artists from \texttt{setlist.fm}. To ensure the resulting TKG remains computationally tractable for current models, we restrict our scope to festivals in Germany\footnote{We provide scripts to mine data for any country/continent.}. 
Germany is suitable due to its large, economically significant, and culturally diverse festival sector \cite{festivalstudie2025}. Moreover, it belongs to the countries with the highest number of entries on \texttt{setlist.fm}, ensuring sufficient data coverage. 

To focus on established festivals, we apply the following filtering criteria: a festival must have occurred at least 5 times and featured at least 30 unique artists overall, counting only editions with a lineup of at least 10 artists.
These thresholds are pragmatic cutoffs to exclude small-scale events.
These filters reduce our dataset to 380 festivals. Given that the earliest festival in our filtered set dates to 1971, we include temporal information from that year onward.

For each artist appearing at these festivals, we extract additional concert data from \texttt{setlist.fm}, including all concert dates, venues, and venue locations. To prevent the TKG from being too large for many TKG forecasting models, and to focus on relevant venues, we exclude venues hosting fewer than five concerts, as such venues typically contribute little structural information while substantially increasing graph sparsity.

Finally, we enrich the artist information with metadata from \texttt{MusicBrainz} via their API. We extract founding (and, if applicable, ending) as well as geographic information, artist types, label relationships, and release history where available.

We started fetching festival information from \texttt{setlist.fm} on Dec 16th, 2025, concert performance data on Feb 9th, 2026, and metainformation from \texttt{MusicBrainz} on Feb 14th, 2026.

Table~\ref{tab:schema} presents the schema for our TKG. The base schema comprises 23 relations organized into four categories: (1) performance information, (2) release information, (3) artist metadata, and (4) label relationships.
We provide this schema as a lightweight ontology in the GitHub repository.



\begin{table}[]
\scriptsize
\centering
\begin{tabular}{p{1.cm}p{0.8cm}p{2.4cm}l}
\toprule
\multicolumn{1}{l}{}                                & Domain   & Relation                               & Range                           \\
\midrule
\multirow{5}{*}{\parbox{1.cm}{Performance \\  Information}} & ARTIST   & \textbf{performs at festival}                 & FESTIVAL                        \\
                                                    & FESTIVAL & happens in venue                     & VENUE                            \\
                                                    & ARTIST   & performs concert at                  & VENUE                            \\
                                                    & VENUE    & venue has location                   & AREA                             \\
                                                    & AREA     & location has country                 & AREA         \\
\midrule
 \multirow{3}{*}{\parbox{1.cm}{Release \\ Information}}                                                     & ARTIST   & releases album                        & RELEASE                          \\
                                                    & ARTIST   & releases EP                           & RELEASE                          \\
					  & RELEASE & has genre 		& GENRE-TAG  \\
\midrule
\multirow{3}{*}{\parbox{1.cm}{Meta \\ Data}}                   & ARTIST   & has begin area                       &AREA          \\
                                                    & ARTIST   & has area                              & AREA                            \\
                                                    & ARTIST   & has type                              & ARTIST-TYPE  \\
\midrule
\multirow{12}{*}{\parbox{1.cm}{Label \\ Relationships }   }              & ARTIST   & label founder                         & LABEL                           \\
                                                    & ARTIST   & recording contract                    & LABEL                           \\
                                                    & ARTIST   & personal label                        & LABEL                            \\
                                                    & ARTIST   & artists and repertoire position at & LABEL                            \\
                                                    & ARTIST   & personal publisher                    & LABEL                           \\
                                                    & ARTIST   & owner                                  & LABEL                           \\
                                                    & ARTIST   & producer position at                 & LABEL                           \\
                                                    & ARTIST   & creative position at                 & LABEL                            \\
                                                    & ARTIST   & executive position at                & LABEL                           \\
                                                    & ARTIST   & engineer position at                 & LABEL                            \\
                                                    & ARTIST   & named after label                    & LABEL                            \\
                                                    & ARTIST   & named after artist                   & LABEL                            \\
\bottomrule
\end{tabular}
\caption{Data schema for the temporal KG. Each temporal triple also comes with a timestamp.}
\vspace{-2em}
\label{tab:schema}
\end{table}

\subsection{Temporal Information}\label{sec:temp}
Our dataset spans from 1971 to 2025. At the time of construction, lineups for 2026 were not yet complete, leading us to conclude the range at 2025.

We employ yearly granularity for the TKG, meaning timestamps represent calendar years. This choice is driven by practical considerations: typical user queries in our use case (e.g., ``Which artists will perform at the Southside Festival next year?'') do not require finer temporal resolution.

Following common approaches in TKG forecasting~\cite{Li2021regcn,gastinger2024tgb}, we encode all temporal information as TKG quadruples, where each triple, e.g., \textit{(Arlo Parks, performs at festival,  Melt! Festival)}, has a timestamp, e.g., 2022. Temporal annotation varies by relation:
\begin{itemize}[leftmargin=*]
\setlength{\itemsep}{0pt}
\setlength{\parskip}{0pt}
\setlength{\parsep}{0pt}
\item {Performance events}: For triples such as \textit{(ARTIST,  performs at festival, FESTIVAL)}, we assign the year of the performance as the timestamp. The same applies to individual concert performances.
\item {Releases}: Release dates (specifically the year) serve as timestamps for album/EP triples and Genre information.
\item {Artist metadata}: Meta-information (e.g., origin area, artist type) is included only when \texttt{MusicBrainz} provides a lifespan start date. For persons, this typically corresponds to birth date and place; for groups, to founding date and place. The relation \textit{has begin area} is timestamped with the artist's start date. In contrast, temporal triples with the relations \textit{has area} and \textit{has type} are valid throughout the artist's active period, defined by the \texttt{MusicBrainz} lifespan start and end dates. If no end date is specified, we assume the artist remains active through the dataset's endpoint.
\item {Labels}: For label relationships, we use the label relation start and end dates from \texttt{MusicBrainz}. When label relation temporal information is unavailable, we default to the artist's active period.
\item {Geographical information}: These triples are typically static. Annotating them at every timestamp would unnecessarily inflate the dataset. Instead, we include them only at timestamps where the corresponding node participates in another relation (e.g., venue location is recorded only in years it hosts a concert).
\end{itemize}

Representing validity as repeated yearly instances rather than time intervals results in a larger graph size. For example, an artist active for 20 years generates 20 distinct instances of the same temporal triple, each with a different timestamp. We accept this trade-off to maintain compatibility with standard TKG forecasting models (see Section~\ref{sec:rel_work}), which typically expect this representation instead of an interval-based one.

\subsection{Dataset Variants}
We present two variants of the TKG, 
each designed to evaluate specific model capabilities.
First, \texttt{concert} is the default variant, which encompasses all performance information, release information, artist metadata, and label relationships described in Table~\ref{tab:schema}. And second, \texttt{concert$_{\texttt{p}}$} is a subset that contains only the quadruples with the relation \textit{performs at festival}.
This variant offers a significantly reduced graph size. It serves two purposes: evaluating models constrained by high memory requirements and testing the extent to which the models can leverage additional information by comparing results from this limited version to the default variant, which contains richer information.

Importantly, the prediction task and the ground truth are consistent across both variants. The quadruples corresponding to \textit{performs at festival} exist identically in \texttt{concert} and \texttt{concert$_{\texttt{p}}$}, ensuring that test set performance is directly comparable regardless of the background knowledge.

In the GitHub repository, we also provide an RDF serialization of the \texttt{concert} dataset. We use RDF-star to encode the temporal triples.

\begin{figure}
    \centering
    \includegraphics[trim={.1cm .1cm 0.2cm .2cm},clip,width=.7\linewidth]{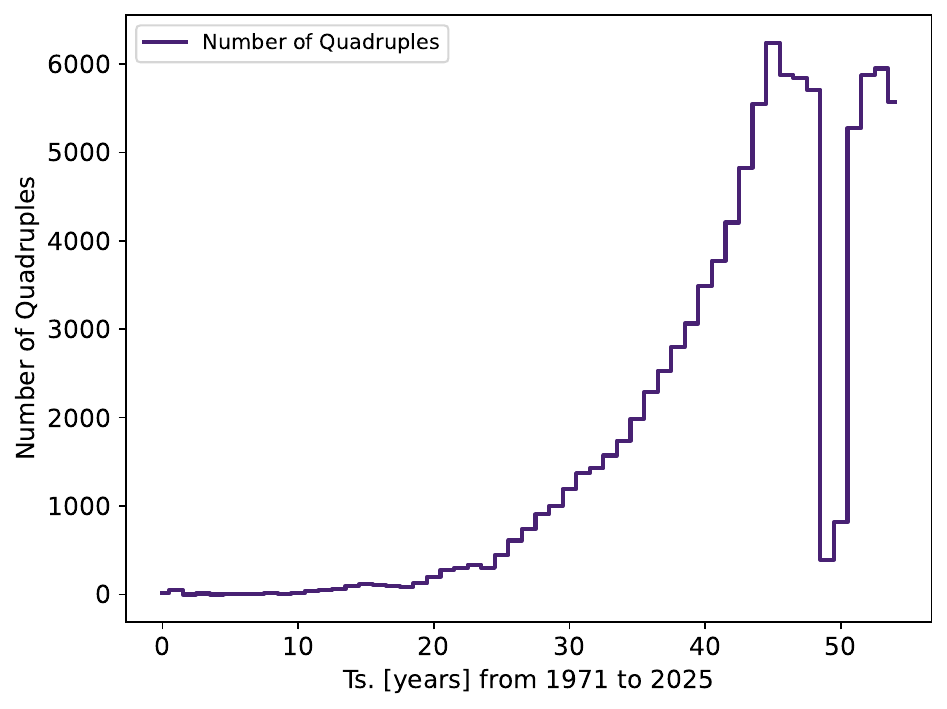}
    \caption{\texttt{concert$_{\texttt{p}}$}, number of quadruples over time.}
    \label{fig:numquad}
\vspace{-1em}
\end{figure}
\subsubsection{Data Splits} We partition the TKG chronologically into train (1971–2023), validation (2024), and test set (2025). 
We chose this split to mitigate the impact of the COVID-19 pandemic, which caused a severe disruption in the occurrence of cultural events. 
Figure~\ref{fig:numquad} depicts the number of quadruples over time for \texttt{concert$_{\texttt{p}}$}, i.e., the number of festival performances.
From 1971 to 2019, it shows a steady growth. A sharp drop is visible during the COVID-19 pandemic (2020–2021). 
Quadruples count returns to nearly pre-pandemic numbers by 2023. 
To prevent this atypical period from dominating the evaluation, we use 2024 as the first validation year. This ensures at least one year of near-typical festival activity (2023) before evaluation.

\newcolumntype{R}[1]{>{\raggedleft\arraybackslash}p{#1}}
\begin{table*}[t]
\centering
\scriptsize
\begin{tabular}{l rrrrrrrR{1.4cm}R{1.4cm}} 
\toprule
Dataset &   $\#$N&   $\#$R&  $\#$Train & $\#$Valid  & $\#$Test & Time  & $\#$Tr/Val/   & DRec [\%] & Rec [\%] \\ 
 &   &   &   &  &   & Int.  &  Te TS  &  \multicolumn{2}{c}{\scriptsize{for rel. of interest}} \\ 
\midrule
  \texttt{concert} &  $278.936$ 	&  $23$ 	& $5.068.265$	& $222.507$	& $214.578$	& 1 y. & 53/1/1       &    \multirow{2}{*}{$2.9$} & \multirow{2}{*}{$29.4$}
 \\
 \texttt{concert$_{\texttt{p}}$} &  $25.399$ 	&  $1$ 	& $83.889$	& $5.949$	& $5.570$	& 1 y. & 51/1/1  & & \\
\bottomrule
\end{tabular}
\caption{Dataset statistics: The number of nodes ($\#$N), relations ($\#$R), train, valid, and test quadruples, the time interval, number of train/valid/test timestamps, as well as Direct Recurrency Degree (DRec) and Recurrency Degree (Rec) for the \textit{performs at festival} relation.}
\vspace{-1em}
\label{tab:dataset}

\end{table*}

\subsubsection{Dataset Statistics} Table~\ref{tab:dataset} summarizes statistics for both variants. We report the Recurrency Degree (Rec), which is defined as the fraction of test quadruples $(s,p,o,\tp)$ for which there exists a $k<\tp$ such that $(s,p,o,k) \in G$, and the \textit{Direct Recurrency Degree (DRec)} which is the fraction of temporal triples $(s,p,o,\tp)$ for which it holds that $(s,p,o,\tp\!-\!1) \in G$~\cite{gastinger2024baselines}. Both metrics are computed exclusively for the quadruples containing the relation 
\textit{performs at festival}. 

For forecasting, a particularly relevant measure is the DRec of \texttt{concert$_{\texttt{p}}$}, which quantifies how often triples to be predicted recur. In contrast to the DRecs of established TKG datasets reported by \citet{gastinger2024baselines} (e.g., DRec of $92.7\%$ for YAGO), the DRec of \texttt{concert$_{\texttt{p}}$} is substantially lower. 
One possible explanation is that consecutive-year re-bookings are relatively uncommon in festival lineups, which leads to fewer directly recurring triples.
This low DRec highlights the difficulty of predicting future lineups from past observations: Models must capture dependencies beyond simple recurrence patterns in order to be successful.


Compared to commonly used TKG forecasting benchmarks, the \texttt{concert} dataset contains a relatively large number of nodes and quadruples and comparatively few timestamps. Thus, individual temporal snapshots are substantially larger than those of most existing datasets. Appendix~\ref{app:dataset} provides a comparison with existing datasets, together with additional dataset statistics, including distributions of relations, genres, and geographic areas.

Constructing and utilizing a TKG for festival lineup prediction involves several design choices and assumptions. We discuss these considerations, along with their limitations, in Appendix~\ref{sec:assumptions}.

\section{Experiments}
\subsection{Evaluation Protocol}
We follow the evaluation protocol discussed in~\citet{gastinger2023eval}. For evaluation, we use and adapt the TGB 2.0 framework~\cite{gastinger2024tgb}.
We evaluate on single-step prediction. We predict entities in both directions, namely $(s, r, ?, t^\star)$ and $(?, r, o, t^\star)$, achieved by introducing inverse relations, where the head and tail of an existing relation are inverted. 
As explained in Section~\ref{sec:task}, we only evaluate on quadruples with the relation \textit{performs at festival} and its inverse.

TKG forecasting is commonly evaluated using time-aware filtered Mean Reciprocal Rank (MRR) and Hits@k~\cite{gastinger2023eval}.
Following the widely adopted filtered ranking protocol introduced by \citet{transe2013} for KG completion, 
MRR and Hits@k are computed per individual test (temporal) triple: for each test triple, the corresponding correct entity is ranked against a set of candidate entities to obtain a reciprocal rank. 
MRR is the average of these reciprocal ranks across all such queries, and Hits@k measures the fraction of these queries for which the correct entity appears in the top-k ranked scores.
If a query has $n$ correct answers, this leads to $n$ test (temporal) triples, each ranked independently. 
The time-aware filtered variant removes other entities known to be correct at the query time from the ranking, to avoid penalizing correct predictions~\cite{gastinger2023eval}.



Our use case differs from common TKG evaluation, where usually each query has only a few correct answers: A single festival may feature over 200 performing artists, meaning each query of the form $(?, \textit{performs at festival}, F, t^\star)$ can have hundreds of correct answers. We argue that MRR and Hits@k yield misleadingly low scores in this setting, and therefore report two additional metrics. 

To illustrate, consider a festival $F$ with 100 ground truth artists, yielding 100 test quadruples of the form $(A_i, \textit{performs at festival}, F, t^\star)$. 
Suppose a model produces a ranking in which every entity at an odd-ranked position (1,3,...,199) is correct, and every entity at an even-ranked position is incorrect. This is not a bad ranking, since half of all predictions are correct.
For the test quadruple of the artist ranked at position 199, the time-aware filter removes the 99 other correct artists from the ranking, leaving this artist at filtered rank 100. 
The MRR across all 100 test quadruples for this festival is then $(1 + 1/2 + 1/3 + ... + 1/100)/100 = 5.2\%$, creating the misleading impression of a near-useless model. 
Yet a ranking in which every other prediction is correct would intuitively warrant a score near $50\%$.

We therefore report two additional metrics:
The \textbf{R-Precision}~\cite{manning2008evaluation} measures how many of the top $R_i$ predicted candidates were correct, where $R_i$ is the number of ground-truth entities for query $i$:
$P_R = \frac{1}{N}\sum_{i=1}^N{\frac{C_{R_i,i}}{R_i}}$
with $C_{R_i,i}$ being the number of correct predictions among the top $R_i$ ranked predictions for query $i$, and $N$ the number of queries.
The \textbf{Normalized 10-Precision} measures how many of the top-10 predictions are correct, normalized by the minimum of $R_i$ and $10$ to account for queries with fewer than 10 correct answers:
 $P_{N10} = \frac{1}{N}\sum_{i=1}^N{\frac{C_{10,i}}{\min(R_i,10)}}$
where $C_{10,i}$ is the number of correct predictions in the top $10$ ranked predictions for query $i$. 

We consider these metrics more reflective of the real-world use: A user asking for festival prediction cares primarily about how many of the model's top-predicted artists (or festivals) actually appear in the lineup (or tour plan), which is exactly what these metrics capture. In our example above, both metrics would result in a score of $0.5$.

\subsection{Methods}
\subsubsection{TKG Forecasting Methods}
We compare the following methods: CountTRuCoLa, TLogic, RE-GCN, CEN, CognTKE, and the Recurrency Baseline, since they represent the best performing methods reported across varying datasets in~\cite{gastinger2025ruc}.

We adapt the rule-based methods to only mine rules whose head relation is among the relations of interest, and all methods to produce predictions only for the queries of interest.
We describe the hyperparameter selection and the hardware used for our experiments in Appendix~\ref{app:expdetails}. 

\subsubsection{LLMs}
We also test whether a TKG and specialized forecasting pipeline are even necessary, or whether zero-shot LLM prompting can perform competitively without any KG data. This is not a comprehensive evaluation of LLM capabilities, but a check on the necessity of the TKG-based approach.

We evaluate four LLMs spanning both open-source and proprietary model families and covering a range of model sizes: Llama 3.1 8B~\cite{meta_llama_3_1_8b}, a quantized Llama 3.3 70B~\cite{ibnzterrell2026metallama33awqint4}, Mixtral 8x7B~\cite{mistral2023mixtral,jiang2024mixtralexperts}, and GPT-4o-mini~\cite{openai2024gpt4omini}. 
All evaluated models have documented knowledge cutoffs no later than late 2023, ensuring that festival lineups from the 2025 test period cannot have been observed during pretraining. Since our goal is to assess whether broadly accessible general-purpose LLMs can compete with specialized TKG forecasting methods, rather than to benchmark the latest frontier models, we focus on models that are relatively efficient, 
are largely locally deployable, and practical to reproduce within typical academic compute budgets.


Each model is prompted to produce a ranked list of artists or festivals for a given query; find full prompts in Appendix~\ref{app:prompts}. We request lists of up to 210 artists and 50 festivals per query, thresholds which slightly exceed the maximum number of artists per festival and festivals per artist in the test period.
To verify that the training cutoff does not substantially bias performance, we evaluate all open-source models on queries from 2023, 2024, and 2025. We find no evidence that the knowledge cutoff substantially degrades 2025 performance, see full study in Appendix~\ref{app:llmcutoff}.
We set the Temperature to 0 throughout to maximize reproducibility. 
Since LLMs produce a candidate list, instead of a full ranking over all entities, it would be unfair to compute MRR and Hits@k. We therefore report only the $P_{N10}$ and $P_{R}$ for LLM results.

We do not provide models with the set of TKG entities, reflecting a realistic deployment scenario in which the full candidate set is unavailable. This has two consequences. First, predictions may include entities absent from the TKG, either due to hallucination or because the entity is not present in the dataset. And second, the predictions must be mapped back to TKG nodes before evaluation.

This mapping is conducted as follows: Since LLMs generate free-text entity names that may differ from the TKG node labels, each predicted name is normalized, e.g., by lowercasing, applying Unicode normalization, and removing punctuation. The resulting string is matched against three reference resources: the artist and entity mapping from the TKG, a worldwide festival mapping from \texttt{setlist.fm}, and an artist mapping containing all artists and their identifiers from \texttt{MusicBrainz}. 
Matching is performed over a set of string variants generated using simple heuristics such as prefix removal (e.g., “the”, “dj”), suffix stripping (e.g., "open air", "festival"), and comma reordering (e.g., “Franklin, Aretha” to “Aretha Franklin”). Predictions that cannot be resolved against any of the three resources are treated as unmatched. To ensure that we did not miss too many correct predictions, we randomly selected 50 unmatched names and tried to match them manually. In none of these cases did we detect a missing match that would have resulted in a correct prediction.

\subsection{Results and Discussion}\label{sec:results}
Table~\ref{tab:results_perrel} reports results per direction: predicting artists \textit{(?, performs at festival, F, 2025)} for given festivals $F$, predicting festivals \textit{(A, performs at festival, ?, 2025)} for given artists $A$, and aggregated over both directions.
RE-GCN, CEN, and CognTKE consistently produced CUDA out-of-memory errors on the full \texttt{concert} dataset even after reducing embedding dimension and batch size, 
and are therefore only evaluated on \texttt{concert$_{\texttt{p}}$}.

\begin{table}[]
\scriptsize 
\scriptsize
\setlength{\tabcolsep}{2.5pt}
\begin{tabular}{@{}lrrrrr|rrrrr@{}}
\toprule
  & \multicolumn{5}{c}{predicting artists} & \multicolumn{5}{c}{predicting festivals}  \\
\midrule
  &    MRR & H1 &  H10 &  P$_R$ &   P$_{N10}$ &  MRR & H1 &  H10 &  P$_R$ &   P$_{N10}$ \\
\midrule
 &   \multicolumn{10}{c}{\texttt{no TKG provided}}   \\ 
GPT4-o-m. & - & - & - & 0.8	&0.7& - & - & - &  4.1	&14.2  \\
Llama 8B & - & - & - &  0.4& 		0.5 & - & - & - & 3.1	&10.2 \\
Llama 70B & - & - & - &  1.3	&1.7& - & - & - & 7.0&17.2  \\
Mixtral 8x7b & - & - & - &  0.6	&0.6 & - & - & - & 5.5	&15.9  \\
\midrule
 &   \multicolumn{10}{c}{\texttt{concert}$_{\texttt{p}}$}   \\ 
CEN & 1.3 & 0.4 & 2.5 & 4.5 & 5.8 & 18.0 & 	9.9	&34.3	&13.1	&33.5 \\
Rec B.  & \underline{2.4} & \textbf{1.2} & 4.2  & 7.1 & 9.4  & 18.3 & 10.9 & 33.8 &  13.4 & 31.9 \\
RE-GCN & 1.4 & 0.3  & 3.0  & 5.1  & 6.5& 16.8 & 8.9  & 32.5  & 11.3  & 30.5\\
CognTKE & \textbf{2.7} & \underline{1.0}  & \textbf{5.5}  & \textbf{9.6}  & \textbf{11.9}&  21.3 & \underline{12.5}  & 40.1  & 14.6 & 34.4\\
CountTR. & 2.3 & {0.9} & 4.3  & 7.5 & 10.2 & \textbf{23.0} & \underline{12.5}  & \textbf{44.3}  & \underline{15.6}  & \textbf{41.4}  \\
TLogic & 0.8 & 0.2  & 1.5 & 3.2 & 3.3 & 11.7 & 6.9 & 22.5  & 9.5  & 21.5 \\
\midrule
 &   \multicolumn{10}{c}{\texttt{concert}}   \\ 
CountTR. &   \underline{2.4}	&{0.9}	& \underline{4.4}	& \underline{7.7}	& \underline{10.5}   &\textbf{23.0} &\textbf{13.5}	& \underline{42.5}	&\textbf{16.5}	& \underline{40.4} \\
TLogic & 0.9	&0.3	&1.5	&3.8	&4.3 & 11.3	&6.5	&22.9	&9.3	&21.6\\
\bottomrule
\end{tabular}
\caption{Test results per direction, all values show $\%$: predicting artists \textit{(?, performs at festival, F, 2025)}, predicting festivals \textit{(A, performs at festival, ?, 2025)}. Best results are \textbf{bold}, second best are \underline{underlined}.}
\label{tab:results_perrel}
\vspace{-2em}
\end{table}

In general, scores are higher for predicting festivals than for predicting artists.
This is expected, as there are significantly less festivals than there are artists, and the set of suitable festivals for a given artist is more constrained, e.g., by genre compatibility, than vice versa.
The best-performing model differs by direction: CognTKE achieves the highest scores for artist prediction, while CountRuCoLa performs best for festival prediction.

\textbf{Does information beyond performance history help?}
CountRuCoLa and TLogic show slight improvements on \texttt{concert} over \texttt{concert}$_{\texttt{p}}$ for artist prediction, but no consistent pattern emerges for festival prediction. 
The enriched data appears to provide value in some cases, but also to introduce noise that current models struggle to cope with.

\textbf{Is there a benefit in using specialized TKG methods over out-of-the-box LLMs?}
Comparing TKG methods to LLMs applied zero-shot, TKG methods show a clear and consistent advantage across both prediction directions.
For artist prediction, the best TKG model CognTKE achieves a $P_{N10}$ of $11.9\%$, compared to only $1.7\%$ for the strongest LLM (Llama~3.3~70B).
This gap holds across all metrics and both prediction directions.
This suggests that the implicit knowledge of a general-purpose LLM cannot replace specialized forecasting methods on a domain-specific TKG.


\textbf{Are the results sufficient for real-world application?}
Artist prediction scores ($P_R$ values of  $8$-$12\%$) are too low to serve as standalone recommendations, but festival prediction is more encouraging:  around $16\%$ of predicted festivals are correct under $P_R$, and $40\%$ within the top-10 predictions.  A realistic deployment could provide model outputs as a ranked shortlist for human review. 
The limited gain from \texttt{concert}$_{\texttt{p}}$ to \texttt{concert} further suggests that current TKG models do not yet capture the full complexity of festival booking. Closing this gap represents a clear direction for future work.

We present additional analyses, including whether prediction quality depends on the activity level of the query subject and the potential benefit of auxiliary shortcut relations, in Appendix~\ref{app:additionalresults}.
\subsection{Discussion of Model Strengths, Limitations, and Identified Failure Modes}\label{sec:dis}
\textbf{LLMs:}
All four models exhibit a strong popularity bias. They frequently predict globally prominent artists rather than artists relevant to the German festival landscape. For example, GPT-4o-mini names Billie Eilish in the top 3 predictions in $47.1\%$ of queries, even though she did not perform at a German festival in 2025.
In addition, LLM-generated rankings contain duplicate and near-duplicate entries (e.g., Llama 3.1 8B predicts Robin Schulz an average of 23 times per query across 24 queries), and show autoregressive effects, where consecutive predictions share properties, such as artist gender or genre, rather than reflecting independent judgments. Detailed examples are provided in Appendix~\ref{app:llmranking}.

\textbf{Embedding Based Methods:}  
All three methods encounter out-of-memory errors on the full \texttt{concert} dataset, likely due to large snapshots and number of entities. RE-GCN and CEN underperform the Recurrency Baseline, consistent with previous observations~\cite{gastinger2024baselines,gastinger2024tgb}. CognTKE shows more promise: it achieves competitive scores even when trained solely on \texttt{concert\textsubscript{p}}, suggesting that research into improved scalability is a worthwhile direction.


\textbf{Rule-Based Methods:} 
CountTRuCoLa is restricted to rules of length~1, thus cannot capture multi-hop dependencies.  
Despite this, it uncovers meaningful connections that reflect plausible real-world connections between artists, venues, festivals, and record labels. E.g., performing at Mayfair Ballroom is predictive of playing at Rock am Ring, and holding a contract with Spinefarm Records is predictive of playing at Wacken Open Air.
TLogic, in contrast, performs substantially below other TKG methods, despite being competitive on standard benchmarks~\cite{gastinger2023eval}.
It only learns cyclic rules without constants, meaning that rule bodies must begin and end with the same entities as rule heads. 
This prevents it from learning rules with specific entities, such as artists who perform at Hurricane will likely perform at Southside Festival.
($<0.2\%$ relative MRR improvement).


\section{Future Work}
Our work opens several directions for future research.
The current TKG could be enriched in multiple ways: Unstructured news data could capture artist availability (tour cancellations, recording schedules). Acoustic attributes could help capture stylistic similarity. Economic factors such as venue capacities and booking budgets reflect practical constraints. 
Moreover, finer-grained geographic information, such as 
inter-city distances, could improve predictions given that German festival organizers place great importance on including local artists~\cite{festivalstudie2025};
future work could account for this via spatial embeddings or proximity-based scores. More broadly, the scalability limitations we observed 
motivate research into more efficient TKG reasoning approaches suited to the large 
real-world event data.
Finally, future work could explore other real-world predictions, such as partial lineup queries (predicting remaining performers given a partially confirmed lineup) and partial schedule queries (predicting additional bookings for a given artist). 
Since partial lineup queries expose some ground-truth information about the target festival, they may substantially change task difficulty; understanding this effect is a useful direction for future work. Both settings would require adapted evaluation protocols.



\bibliography{refs}

\clearpage
\appendix

\section{Appendix}
\definecolor{viridisblue}{HTML}{1f998a} 
\definecolor{viridis2}{RGB}{44, 113, 142}
\subsection{Assumptions and Limitations}\label{sec:assumptions}
Constructing and utilizing this TKG for festival lineup prediction introduces several challenges and is based on certain assumptions:

\textbf{Assumption of Completeness:} The forecasting models and evaluation protocols operate under the assumption that the TKG is sufficiently complete regarding the festival lineups. Both setlist.fm and MusicBrainz are crowdsourced and thus are prone to coverage gaps. We performed sanity checks by checking a sample of lineups against official festival websites. These checks revealed no major omission. 

\textbf{Dataset Scope:} The current schema is limited to performances, releases, metadata, and label information. It excluded additional contextual information that human experts might leverage, such as unstructured data from news and announcements regarding, e.g., artist breaks, recording schedules, or tour cancellations, music-specific details like instrumentation or acoustic characteristics of the music, and economic factors like ticket prices and venue capacities. 
Incorporating such multimodal data could significantly enhance predictive performance but lies outside the scope of this work.

\textbf{Temporal Dynamics and Announcement Waves:} Festival lineups are typically not released at once, but in waves over several months leading up to the event. 
This dynamic release schedule introduces two challenges for our evaluation:
The exact moment when a festival's lineup is considered complete is ambiguous on setlist.fm. For this reason, we excluded the year 2026 from our evaluation, even though partial lineups are available at the time of writing. 
And second, comparing against models with access to real-time web search, e.g., Large Language Models (LLMs), is complicated by the fluid information availability. For instance, asking an LLM at the end of 2025 to forecast lineups for 2026 festivals risks test set leakage, as partial lineup data may already exist online. 

\textbf{Scalability and Memory Constraints:} As illustrated in Figure~\ref{fig:datacomp}, the resulting TKG is large when compared to standard datasets in terms of node and quadruple count, and especially snapshot size. 
This presents a practical problem: many existing TKG forecasting models encounter out-of-memory errors when processing TKG of this size~\cite{gastinger2024tgb}. We already restricted the dataset to festivals located in Germany. Expanding the TKG to a global scale would increase its size and likely make current state-of-the-art models inapplicable without architectural modifications or sampling strategies. 
While testing on a global festival TKG was beyond the scope of this work, we provide the code to enable the construction of such expanded TKGs for future research.


\subsection{Additional Dataset Details}\label{app:dataset}
\begin{figure}
    \centering
    \includegraphics[width=0.8\linewidth]{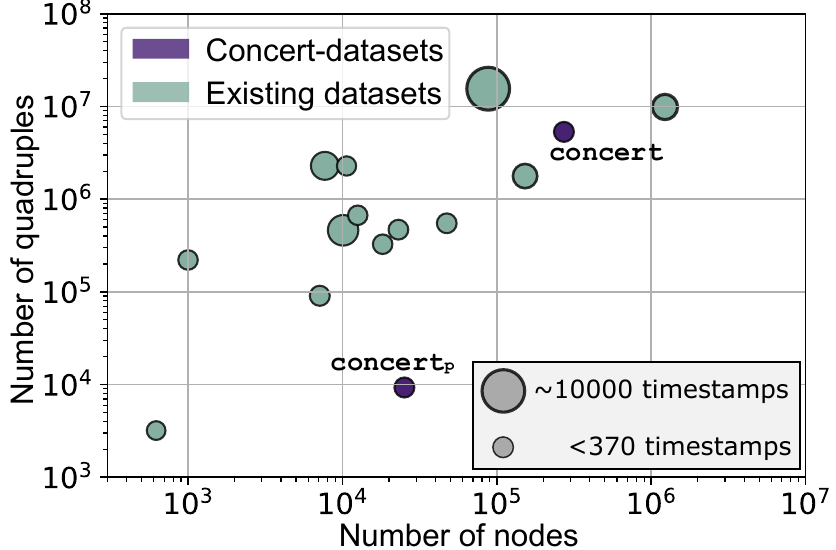}
    \caption{Comparison of \texttt{concert} dataset statistics to existing TKG datasets. Figure adapted from~\cite{gastinger2024tgb}.}
    \label{fig:datacomp}
\end{figure}

Figure~\ref{fig:datacomp} contextualizes the size of our dataset by comparing it to datasets commonly used for evaluating TKG forecasting methods~\cite{gastinger2024tgb}. 
The dataset \texttt{concert} contains a comparatively high number of nodes and quadruples. At the same time, despite its overall size, it comprises relatively few timestamps. This implies that individual temporal snapshots are larger than in most existing datasets.  \texttt{concert$_{\texttt{p}}$} exhibits a comparatively small number of quadruples since it is a subset containing only the quadruples of one relation.

\begin{figure*}
    \centering
    \begin{subfigure}[t]{0.45\linewidth}
         \includegraphics[trim={.1cm .1cm 0.2cm .2cm},clip,width=\linewidth]{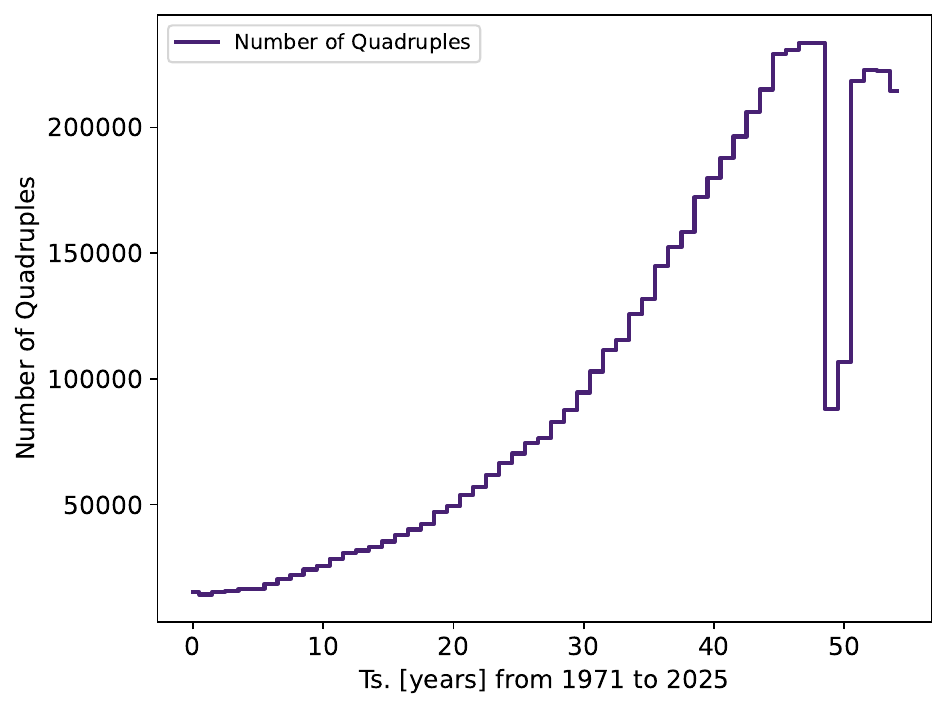}
\caption{\texttt{concert}}
    \end{subfigure}
     \hfill
    \begin{subfigure}[t]{0.45\linewidth}
        \includegraphics[trim={.1cm .1cm 0.2cm .2cm},clip,width=\linewidth]{figs/num_edges_discretized_55_tkgl-concertperformanceonly2.pdf}
	\caption{\texttt{concert$_{\texttt{p}}$}}
 \end{subfigure}    
    \begin{subfigure}[t]{0.45\linewidth}
         \includegraphics[trim={.1cm .1cm 0.2cm .2cm},clip,width=\linewidth]{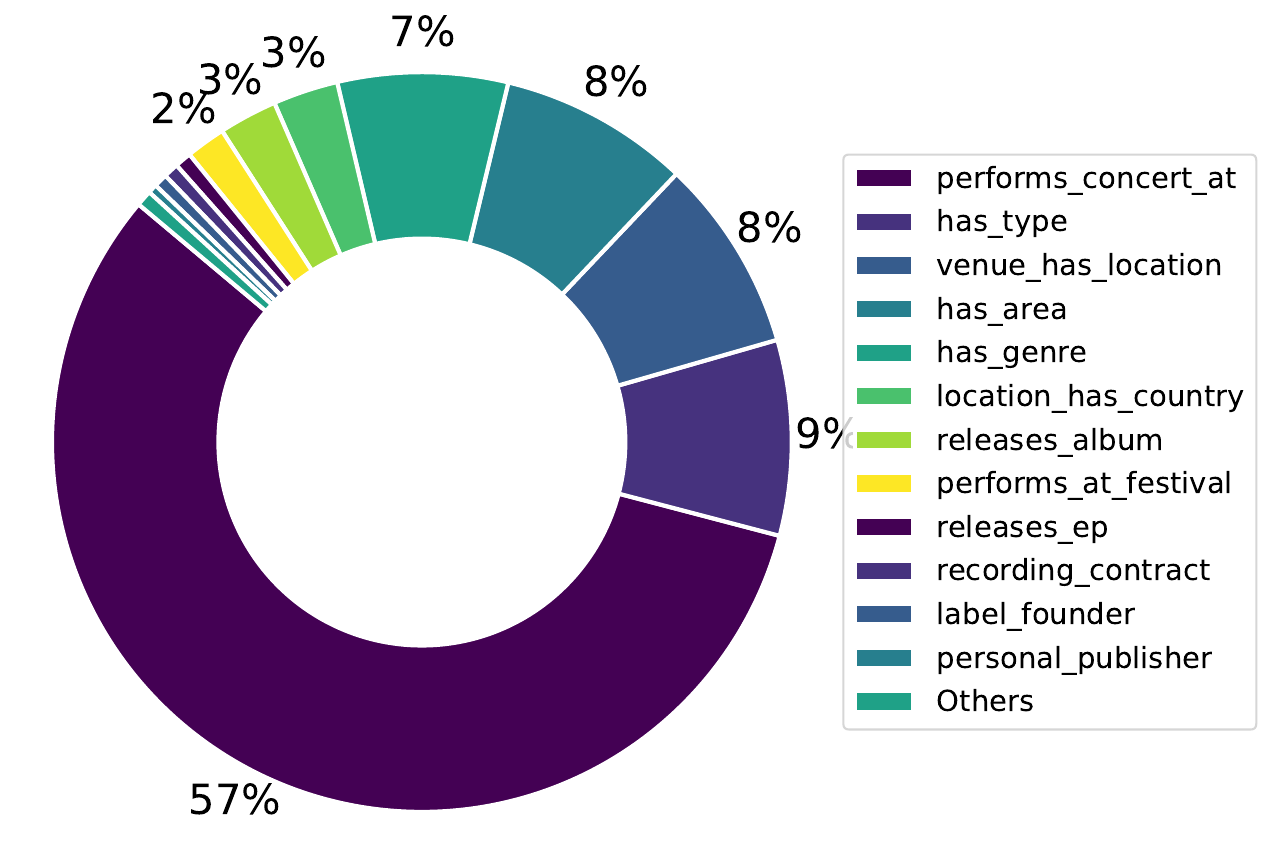}
\caption{Relations}
    \end{subfigure}
     \hfill
    \begin{subfigure}[t]{0.41\linewidth}
         \includegraphics[trim={.1cm .1cm 0.2cm 0cm},clip,width=\linewidth]{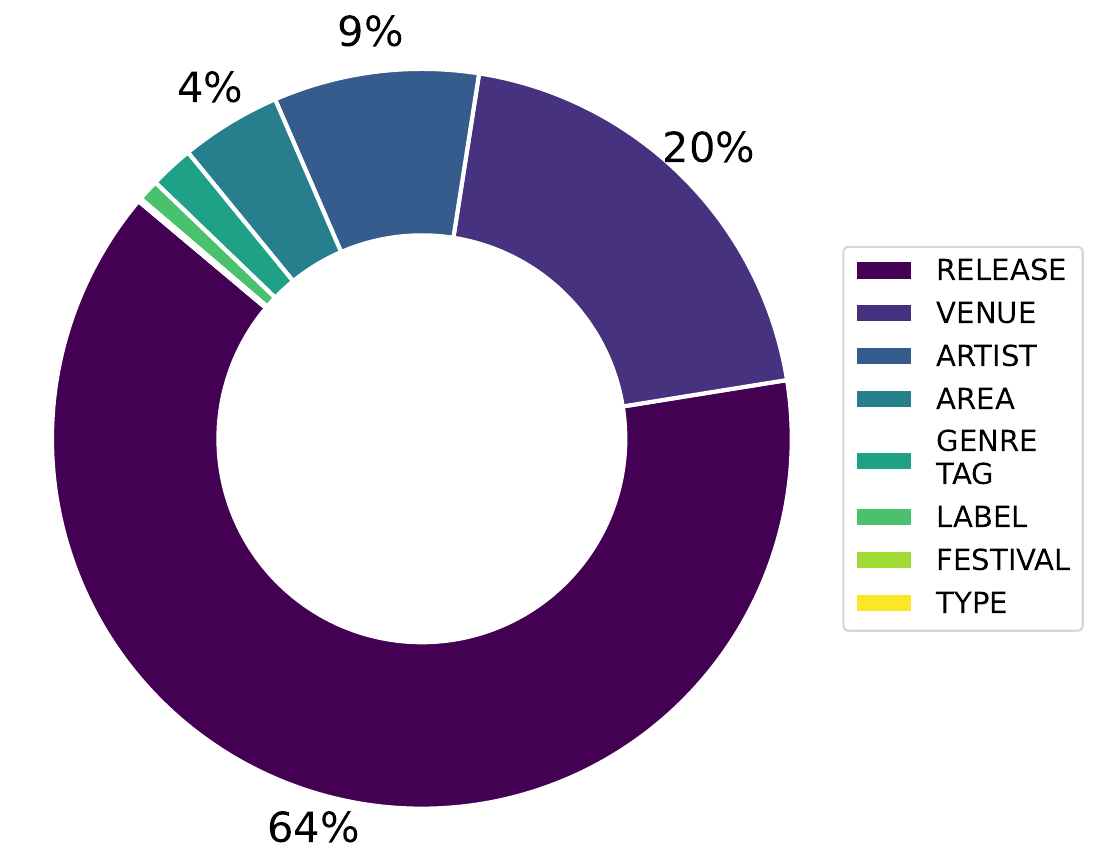}
\caption{Node types}
    \end{subfigure}
    \begin{subfigure}[t]{0.45\linewidth}
        \includegraphics[trim={.1cm .1cm 0.2cm .2cm},clip,width=\linewidth]{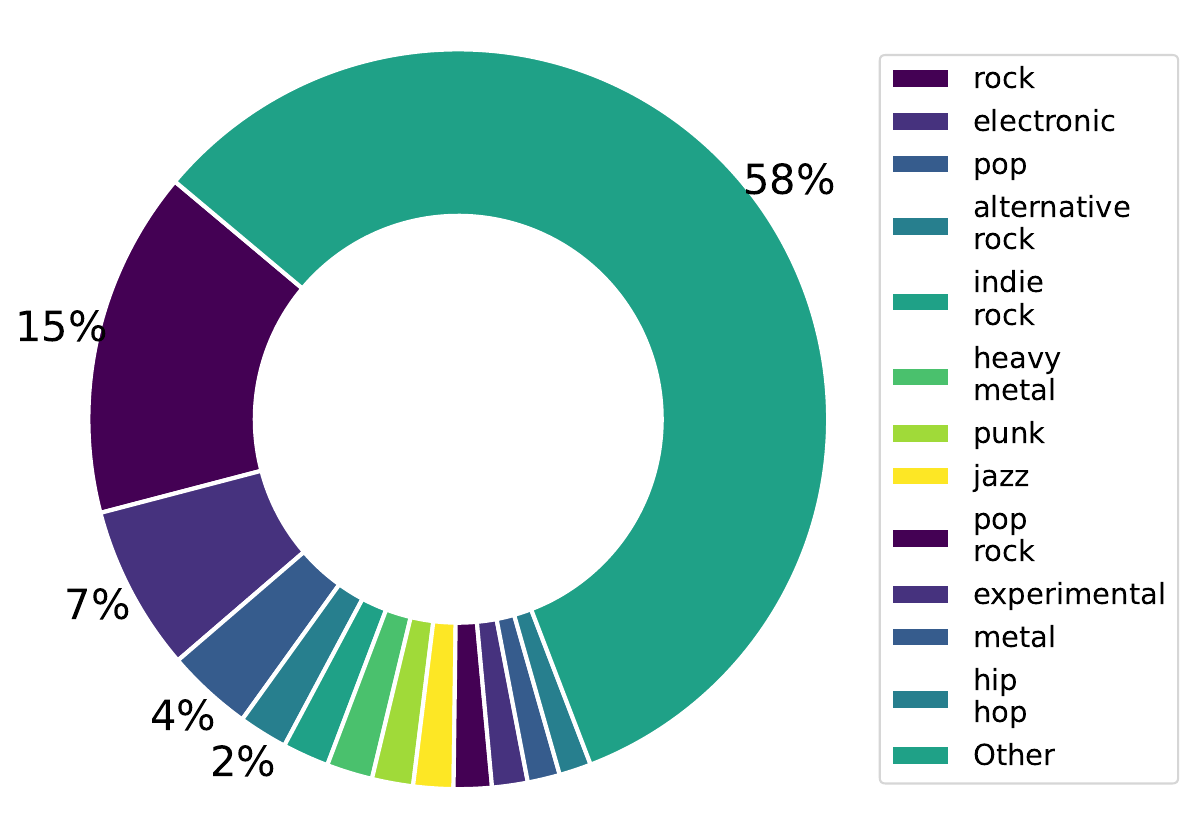}
	\caption{Release Genres}
 \end{subfigure} 
  \hfill
     \begin{subfigure}[t]{0.45\linewidth}
        \includegraphics[trim={.1cm .1cm 0.2cm .2cm},clip,width=\linewidth]{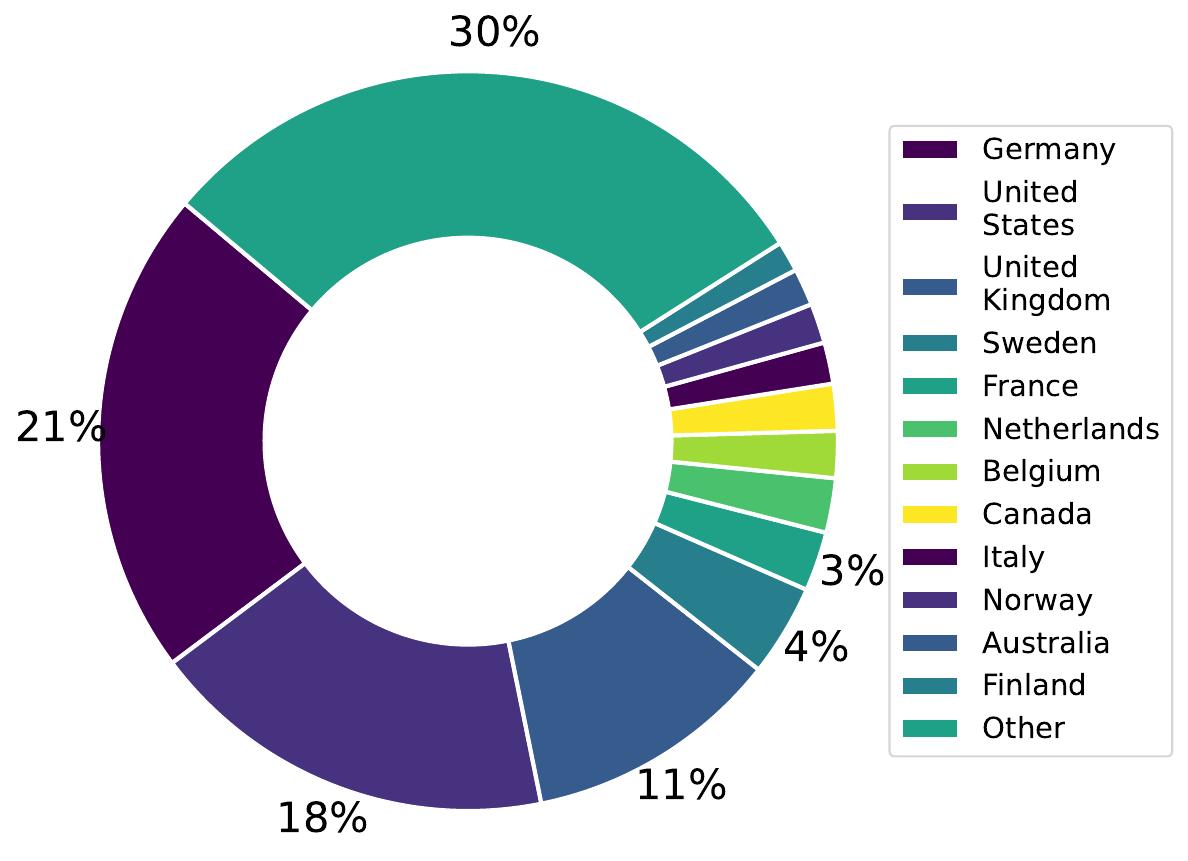}
	\caption{Artist Areas}
 \end{subfigure} 
    \caption{Top: Number of quadruples over time for \texttt{concert} and \texttt{concert$_{\texttt{p}}$}.
    Bottom: Pie charts representing distribution of relations, node types, release genres, and artist areas for\texttt{concert}.}
    \label{fig:datastats}
    \vspace{-16pt}
\end{figure*}

Figure~\ref{fig:datastats} illustrates additional dataset statistics.
Figures~\ref{fig:datastats}(a) and~(b) depict the number of quadruples over time.
Both datasets show steady growth from 1971 to 2019, followed by a sharp drop during the COVID-19 pandemic (2020--2021), which is particularly pronounced in  \texttt{concert$_{\texttt{p}}$}. Quadruple counts recover to near pre-pandemic levels by 2023.

The distribution of relations (Figure~\ref{fig:datastats}(c)) in \texttt{concert} reveals that concert performances dominate the TKG with $57\%$ of all quadruples (relation \textit{performs concert at}). 
The high occurrence of the \textit{has type} relation is caused by our temporal encoding strategy: as detailed in Section~\ref{sec:temp}, this information is replicated for every year of an artist's active period, inflating the count of these specific triples. 

Regarding the distribution of node types (Figure~\ref{fig:datastats}(d)), nodes that represent releases constitute the largest part. This is intuitive: while a single venue hosts a multitude of artists, a single artist may release many albums or EPs, each represented as a distinct node. 
Figure~\ref{fig:datastats}(e) shows the 12 most common release genres: Rock, Electronic, and Pop are the dominant categories, reflecting the mainstream focus of the festivals in the dataset. Figure~\ref{fig:datastats}(f) visualizes the geographic distribution of artist areas: although all festivals take place in Germany, the artists 
are predominantly from Europe and North America, with Germany, the United States, and the United Kingdom contributing the largest shares.

\subsubsection{Selected Dataset Facts}
Beyond aggregate statistics, the dataset enables extracting interesting facts regarding artists and festivals, for example:
\begin{itemize}[leftmargin=*]
\setlength{\itemsep}{0pt}
\setlength{\parskip}{0pt}
\setlength{\parsep}{0pt}
\item The three largest festival editions in terms of artist line-up size are Reeperbahn Festival 2024 (390 artists), Wacken Open Air 2023 (228 artists), and Wave-Gotik-Treffen 2024 (185 artists).  
\item The artists who performed at the highest number of festivals in Germany overall are Donots (63 festivals), Emil Bulls (58 festivals), and Itchy (56 festivals).  
\item The artists who performed at the highest number of festivals in a single year in Germany are Kraftklub (2019; 19 festivals), AnnenMayKantereit (2014; 18 festivals), and Gurr (2016; 17 festivals).  
\item The artists with the longest festival history in Germany (highest number of festival years) are Guru Guru (31 years; first: 1971, last: 2025), Subway to Sally (30 years; first: 1992, last: 2025), and Project Pitchfork (30 years; first: 1993, last: 2025).  \item The venues hosting the highest number of distinct artists are Grote Zaal (2420 artists), Markthalle Hamburg (2310 artists), and Ancienne Belgique (2225 artists).  
\item The most common origin areas of artists are London (457 artists), Berlin (323 artists), and Los Angeles (267 artists).  
\item The countries with the highest number of artists (relation \textit{has area}) are Germany (3424 artists), the United States (2877 artists), and the United Kingdom (1799 artists).  
\item Among artists with available type information (relation \textit{has type}), $73.2\%$ are classified as Group and $26.7\%$ as Person. Additional types (in negligible numbers) include Orchestra, Choir, Other, and Character.  
\end{itemize}

\subsection{Experimental Details}\label{app:expdetails}

\subsubsection{Details on TKG method Experiments}
Hardware Information: The methods RE-GCN, CEN, and CognTKE require execution on a GPU. We run them on an NVIDIA RTX 6000 Ada (48GB VRAM) with 500GB system RAM. The other experiments were run on a SLURM-managed CPU cluster. Nodes featured Intel Xeon (E5-2640 v2/v3/v4, Silver 4114) and AMD EPYC (7413, 7713P, 9474F) processors, with up to 96 cores and 1.5 TB RAM per node. We restricted parallelization to a maximum of 20 concurrent threads and limited available RAM to 500 GB.

For each method, we select the hyperparameters from the ranges specified in the original papers and code repositories. We conduct grid search on the validation set. Table~\ref{tab:hyperparams} reports the values of selected hyperparameters for each method. 
\begin{table*}[]
    \centering
    \begin{tabular}{l|p{10.2cm}}
    \toprule
    Method & Hyperparameters selected \\
    \midrule
    RE-GCN   &  n-layers $= 2$, train-history-len$ = 4$, test-history-len$ = 4$ \\
    CEN     & start-history-len$=1$ dropout$=0.2$ n-layers$=2$ \\
    CognTKE & n-layers $= 3$, window-size $=15$, dropout=$0.25$ (default hyperparameters, since no ranges or tuning procedure were specified, and results on all datasets in original paper were reported on the same hyperparameter values) \\
    CountTRuCoLa & $\mathcal{P}=30$, $\mathcal{P}_{f\text{-rules}}=0$, $\mathcal{C}=True$, $\mathcal{W}=20$, $\mathcal{M}=50$, $\mathcal{Z}=0.01$, $\mathcal{H}=10$, $\mathcal{D}=0.4$, $\mathcal{C}_{Xcount}=3$; for \texttt{concert}$_{\texttt{with shortcuts}}$ we set $\mathcal{C}_{Xcount}=10$, due to larger TKG size.\\    
    TLogic & num-walks$=10$, transition-distr$=exp$, window$=30$, top-k$=10$, $\alpha=1$ , $\lambda=0.01$, rule-length$=[1,2,3]$ \\
    Rec. & No Hyperparameters \\
    \bottomrule
    \end{tabular}
    \caption{Hyperparameter Values selected for each method. }
    \label{tab:hyperparams}
    \vspace{1em}
\end{table*}


\begin{figure*}[t!]
\centering
\begin{Verbatim}
You are estimating likely performers for the \{name\} \{year\} 
in \{city\}, \{country\}.
Task:
Generate a ranked list of 210 artists most likely to perform.
Output requirements:
- Provide exactly 210 artist names
- Rank them from 1 (most likely) to 210 (least likely)
- Output ONLY the ranked list
- Format: each line should be "1. Artist Name", 
  "2. Artist Name", etc.
- Do NOT include explanations, notes, or extra text
If uncertain, still provide your best estimate based on 
the criteria above.
\end{Verbatim}

\caption{LLM Prompts for receiving artists playing at a festival.}
\label{fig:prompt1}
\begin{Verbatim}
You are estimating which festivals in Germany an artist 
named "\{name\}" is most likely to perform at in \{year\}.
Task:
Generate a ranked list of 50 German festivals 
where "\{name\}" is most likely to perform in \{year\}.
Output requirements:
- Provide exactly 50 festival names
- Rank them from 1 (most likely) to 50 (least likely)
- Output ONLY the ranked list
- Format: each line should be "1. Festival Name", 
  "2. Festival Name", etc.
- Do NOT include explanations or any extra text
If uncertain, still provide your best estimate based on 
the criteria above.
\end{Verbatim}
\caption{LLM Prompts for receiving festivals where an artist will play.}
\label{fig:prompt2}
\end{figure*}

\subsubsection{Details on LLM Experiments}\label{app:prompts}
\paragraph{LLM Prompts:}
The prompts used for the LLMs were evaluated and tuned on an iterative approach, generating two prompts for artists and festivals repeatedly\footnote{The prompts, as well as input and evaluation code are also available on GitHub: \url{https://github.com/JuliaGast/festival-lineup-tkg/tree/main/forecasting/llm}.}.

For receiving artists playing at a festival, the LLMs were prompted like in Figure~\ref{fig:prompt1}.

For receiving festivals where an artist will play, the LLMs were prompted like in Figure~\ref{fig:prompt2}. 


\subsection{Additional Experiments and Results}\label{app:additionalresults}
\subsubsection{Results by Subject Activity Level}
\begin{figure*}[t]
    \centering
    \vspace{-0.5em}
    \includegraphics[width=0.9\linewidth]{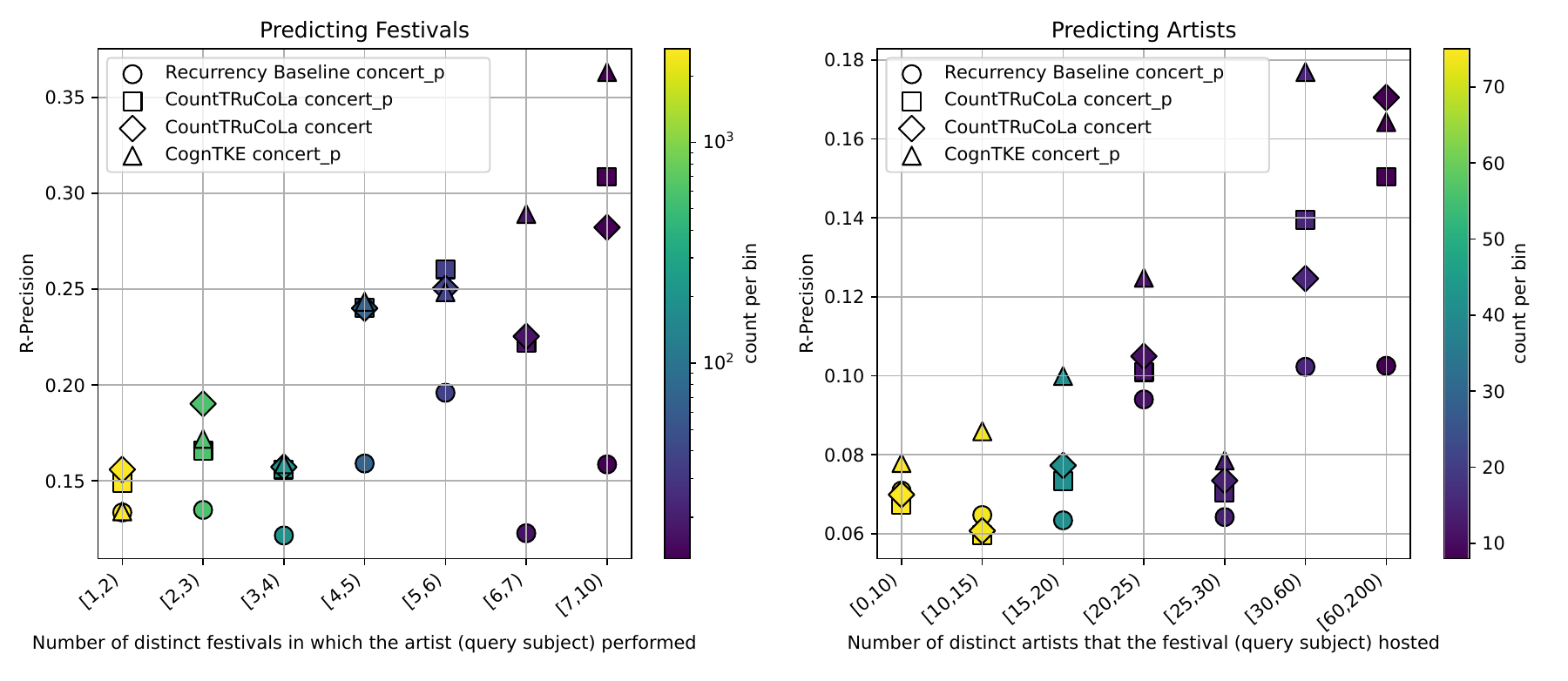}
    \caption{Mean $P_R$ binned by artist activity (number of distinct festivals per year, left) and festival activity (number of distinct artists hosted per year, right) for the four best-performing methods. Colors indicate counts per bin.}
    \label{fig:distinct_artists}
\end{figure*}

To investigate whether prediction quality depends on the activity level of the query subject, we analyze $P_R$ across the number of distinct festivals an artist performed in, and the number of distinct artists a festival hosted (Figure~\ref{fig:distinct_artists}).
A clear positive trend is visible in both directions: prediction quality increases with activity level. 
The Recurrency Baseline performs comparably to other methods for infrequent subjects but falls increasingly behind as activity grows.
This suggests that for infrequent query subjects, all methods are limited to simple distribution heuristics, and the more expressive models offer little advantage.
For more active artists and  festivals, richer temporal and relational patterns become available, which specialized TKG methods are better able to exploit.
Finally, for less active artists, CountRuCoLa benefits more from the full \texttt{concert} dataset. This fits the intuition that additional information like venue patterns adds value when performance histories are sparse.

\subsection{Addition of shortcut relations}
To address the challenge that multi-hop inference across distant nodes may pose for certain TKG models, in this study, we introduce auxiliary shortcut relations under the category \textit{Additionally Added Shortcuts}. These relations do not introduce new information but provide direct edges between artists and attributes (e.g., countries where they perform, genres of their releases) thereby offering an alternative path to information that would otherwise require traversing multiple hops. While this creates redundancy and increases graph size, it may benefit models with limited hop capacity.

We treat shortcut relations as optional input and conduct experiments under two configurations: with and without shortcuts.
Table~\ref{tab:schema_ext} details the schema extensions, and Table~\ref{tab:dataset_ext} reports dataset statistics for \texttt{concert$_{\texttt{shortcuts}}$} alongside the default \texttt{concert} and the reduced \texttt{concert$_{\texttt{p}}$}. 
While \texttt{concert} and \texttt{concert$_{\texttt{shortcuts}}$} share identical node counts, due to the additional relations that we injected in \texttt{concert$_{\texttt{shortcuts}}$}, the number of quadruples increases. 
Importantly, the prediction task as well as the ground truth, are consistent across both variant: the \textit{performs at festival} quadruples exist unchanged in all three datasets, ensuring that test-set results are directly comparable.

Table~\ref{tab:results_shortcuts} reports results for CountTRuCoLa and TLogic on both dataset variants, \texttt{concert$_{\texttt{shortcuts}}$} and \texttt{concert}. The remaining methods could not be evaluated due to out-of-memory errors. 
No consistent improvement is observed when adding shortcut relations, and we therefore conclude that, at least under the current model capabilities, they do not provide a measurable benefit.

\begin{table*}[]
\centering
\begin{tabular}{p{2.5cm}lll}
\toprule
\multicolumn{1}{l}{}                                & Domain   & Relation                               & Range                           \\
\midrule
\multirow{6}{*}{\parbox{2.5cm}{ Additionally \\ Added \\ Shortcuts}}     & ARTIST   & releases album of genre             & GENRE-TAG                      \\
                                                    & ARTIST   & releases ep of genre                & GENRE-TAG                      \\
                                                    & ARTIST   & performs festival in country        & AREA \\ 
                                                    & ARTIST   & performs festival in location       & AREA                             \\
                                                    & ARTIST   & performs concert in country         & AREA\\ 
                                                    & ARTIST   & performs concert in location        & AREA                             \\
\bottomrule
\end{tabular}
\caption{Data schema, extended triples, for the temporal KG. Each temporal triple also comes with a timestamp.}
\label{tab:schema_ext}
\end{table*}

\newcolumntype{R}[1]{>{\raggedleft\arraybackslash}p{#1}}
\begin{table*}[]
\centering
\scriptsize
\begin{tabularx}{\linewidth}{p{2.7cm} R{1.1cm} R{0.5cm} R{1.3cm} R{1.08cm} R{1.08cm} R{0.7cm} R{1.4cm} R{0.7cm} R{0.7cm}} 
\toprule
Dataset &   $\#$N&   $\#$R&  $\#$Train & $\#$Valid  & $\#$Test & Time Int. & $\#$Tr/Val/ Te TS   & DRec [\%] & Rec [\%] \\ 
\midrule
\texttt{concert} &  $278.936$ 	&  $23$ 	& $5.068.265$	& $222.507$	& $214.578$	& 1 y. & 53/1/1          & $26.9$ & $44.2$
 \\
\texttt{concert$_{\texttt{shortcuts}}$}&  $278.936$ 	&  $29$ 	& $8.927.809$	& $423.314$	& $409.911$	& 1 y. & 53/1/1          & $25.0$ & $53.1$
 \\
\texttt{concert$_{\texttt{p}}$} &  $25.399$ 	&  $1$ 	& $83.889$	& $5.949$	& $5.570$	& 1 y. & 51/1/1  & $2.9$ & $29.4$ \\
\bottomrule
\end{tabularx}
\caption{Dataset statistics for all three dataset variants: The number of nodes ($\#$N), relations ($\#$R), train, valid, and test quadruples, the time interval, number of train/valid/test timestamps, Direct Recurrency Degree (DRec), and Recurrency Degree (Rec).}
\label{tab:dataset_ext}
\end{table*}

\begin{table*}[]
\centering
\scriptsize  
\begin{tabularx}{.9\linewidth}{l@{\hspace{-1pt}} rrrrr@{\hspace{2pt}| \hspace{0pt}} rrrrr@{\hspace{2pt}| \hspace{0pt}} rrrrr}
\toprule
  & \multicolumn{5}{c@{\hspace{2pt}| \hspace{0pt}} }{predicting artists} & \multicolumn{5}{c@{\hspace{2pt}| \hspace{0pt}} }{predicting festivals} & \multicolumn{5}{c}{both directions} \\
\midrule
  &    MRR & H1 &  H10 &  P$_R$ &   P$_{N10}$ &  MRR & H1 &  H10 &  P$_R$ &   P$_{N10}$ &  MRR & H1 &  H10 &  P$_R$ &  P$_{N10}$\\
\midrule
 &   \multicolumn{14}{c}{\texttt{concert}}   \\ 
CountTRuCoLa  &  {2.4}	&{0.9}	&{4.4}	&{7.7}	&\textbf{{10.5}}   &\textbf{{23.0}} &{13.5}	&\textbf{{42.5}}	&{16.5}	&\textbf{{40.4}} & \textbf{{12.7}}&{7.2}&\textbf{{23.5}	}&{16.0}&\textbf{{38.5}} \\
TLogic & 0.9	&0.3	&1.5	&3.8	&4.3 & 11.3	&6.5	&22.9	&9.3	&21.6& 6.0	&3.2	&12.2	&8.7	&20.6\\
\midrule
& \multicolumn{14}{c}{\texttt{concert}$_{\texttt{with shortcuts}}$}  \\
CountTRuCoLa  &\textbf{ 2.5}	&\textbf{1.0	}&\textbf{4.5}	&\textbf{8.2	}&\textbf{10.5} & 22.5&	\textbf{13.6	}&40.9	&\textbf{16.7	}&39.0 &12.5	&\textbf{7.3	}&22.7	&\textbf{16.2	}&37.3\\
TLogic & 0.9	&0.1	&1.5	&3.1	&2.9 &11.5	&6.6	&22.8	&9.5	&21.4 & 6.1	&3.3	&12.3	&8.8	&20.4\\ 
\bottomrule
\end{tabularx}
\caption{Test results per direction: predicting artists \textit{(?, performs at festival, F, 2025)}, predicting festivals \textit{(A, performs at festival, ?, 2025)}, and predicting both directions, i.e. aggregated over all test queries. Best results are \textbf{bold}. Comparing the default {\texttt{concert}} dataset to the extended {\texttt{concert}$_{\texttt{with shortcuts}}$} dataset version.}
\label{tab:results_shortcuts}
\end{table*}


\subsubsection{LLM Ranking Quality}\label{app:llmranking}

\paragraph{Detailed Examples on LLM failures: }
All evaluated LLMs exhibit a tendency to favor highly popular artists regardless of their relevance to the German festival landscape. Llama 3.1 8B predicts Taylor Swift among the top three candidates in $23.9\%$ of queries, Mixtral 8$\times$7B in $36.1\%$, and GPT-4o-mini predicts Billie Eilish in $47.1\%$ of queries, although neither artist performed at a German festival in 2025. Llama 3.3 70B instead tends to predict prominent German artists, such as Rammstein ($48.3\%$) and Die Toten Hosen ($23.5\%$), but these predictions are almost always incorrect; only Die Toten Hosen appearing at Jamel rockt den Förster in 2025 was a valid prediction.

The generated rankings also frequently contain duplicate and near-duplicate predictions. For example, Llama 3.1 8B predicts Robin Schulz an average of 23 times per query across 24 queries. Mixtral 8$\times$7B produces near-duplicates such as ``Taylor Swift'' and ``Taylor Swift \& Ed Sheeran'' within the same ranking.

\paragraph{Ranking Example (Autoregressive Bias):}
LLM-generated rankings exhibit autoregressive bias. Since predictions are generated sequentially, previous tokens can influence subsequent outputs, leading to rankings where neighboring predictions share properties such as artist gender or genre rather than representing independent candidate evaluations.

In Figure~\ref{fig:rankingllama}, we show a ranking produced by Llama 3.3 70B when asking who will perform at the \textit{Keep It True Festival} in 2025. Female artists and female-fronted bands are marked in \textcolor{viridisblue}{green}. Rather than being spread throughout the ranking as knowledge of the festival would suggest, these acts form a contiguous block. Artist gender is not a meaningful criterion for ranking bands at a traditional heavy metal festival, so the observed clustering exemplifies the auto-regressive bias described in Section~\ref{sec:dis}: once the model generates one female-fronted act, subsequent token probabilities shift toward similar acts. 

\begin{figure*}[t]
\centering

\begin{minipage}[t]{0.45\textwidth}
\begin{Verbatim}[commandchars=\\\{\}]
1. Saxon 
2. Judas Priest
3. Iron Maiden
4. Scorpions
5. Helloween
6. Blind Guardian
7. Accept
8. U.D.O.
9. Sabaton
10. Powerwolf
11. Rhapsody of Fire
12. Hammerfall
13. Stratovarius
14. Sonata Arctica
15. Dragonforce
16. Alestorm
17. Korpiklaani
\textcolor{viridisblue}{18. Eluveitie}
19. Ensiferum
20. Finntroll
\end{Verbatim}
\end{minipage}
\hfill
\begin{minipage}[t]{0.45\textwidth}

\begin{Verbatim}[commandchars=\\\{\}]
21. Turisas
22. Sabaton
23. Amon Amarth
24. Children of Bodom
25. In Flame s
26. Dark Tranquillity
27. At the Gates
\textcolor{viridisblue}{28. Arch Enemy}
\textcolor{viridisblue}{29. Nightwish}
\textcolor{viridisblue}{30. Epica}
\textcolor{viridisblue}{31. Within Temptation}
\textcolor{viridisblue}{32. Delain}
\textcolor{viridisblue}{33. Leaves' Eyes}
\textcolor{viridisblue}{34. Xandria}
\textcolor{viridisblue}{35. Lacuna Coil}
\textcolor{viridisblue}{36. Evanescence}
37. Kamelot
38. Symphony X
39. Evergrey
40. Threshold
\end{Verbatim}
\end{minipage}
\caption{A ranking produced by Llama 3.3 70B when asking who will perform at the \textit{Keep It True Festival} in 2025. Female artists and female-fronted bands are marked in \textcolor{viridisblue}{green}.}
\label{fig:rankingllama}
\end{figure*}

\subsubsection{Effect of LLM Knowledge Cutoff}\label{app:llmcutoff}
Since the knowledge cutoff of the evaluated LLMs falls around late 2023, queries about 2025 festival lineups may fall outside their reliable knowledge window. To assess whether this introduces a systematic bias, we compare model performance across three years, 2023, 2024, and 2025, using a random subset of 50 artists and 50 festivals sampled from quadruples occurring in all three years. To limit expenses, these tests were only conducted on the three open-source models. 

\begin{figure*}
    \centering
    \begin{subfigure}[t]{0.75\linewidth}
         \includegraphics[width=\linewidth]{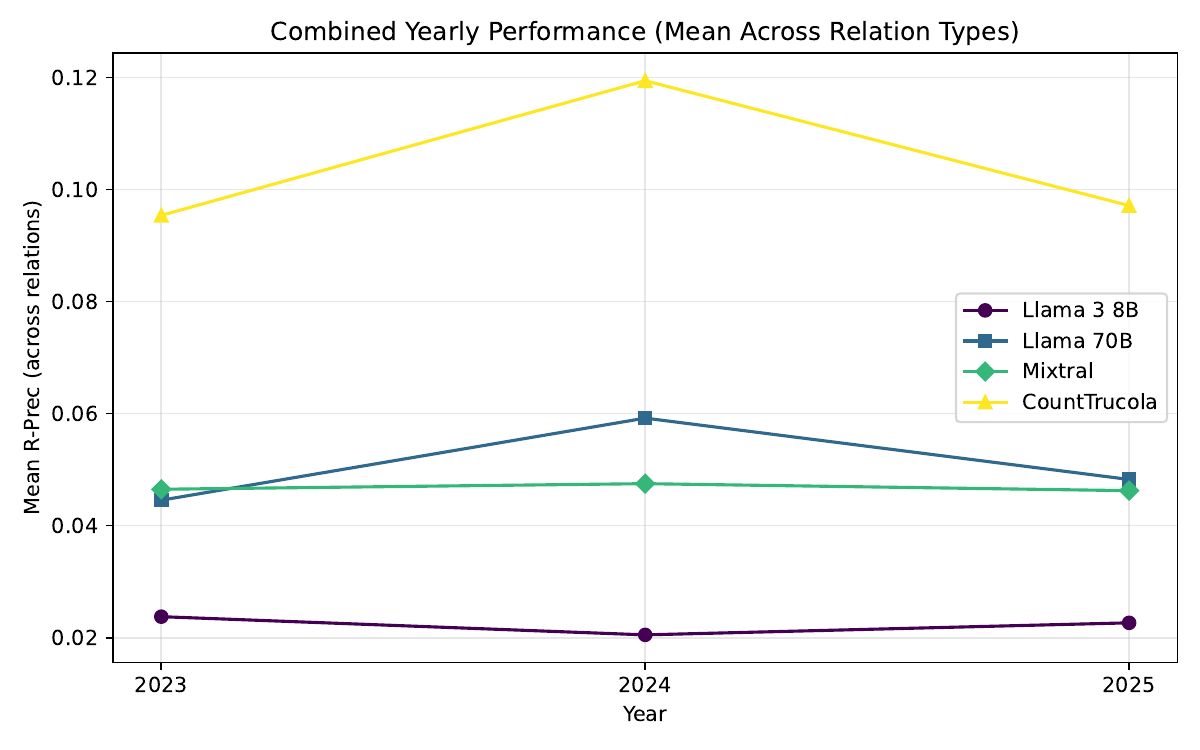}
    \end{subfigure}
     \hfill
    \begin{subfigure}[t]{0.75\linewidth}
        \includegraphics[width=\linewidth]{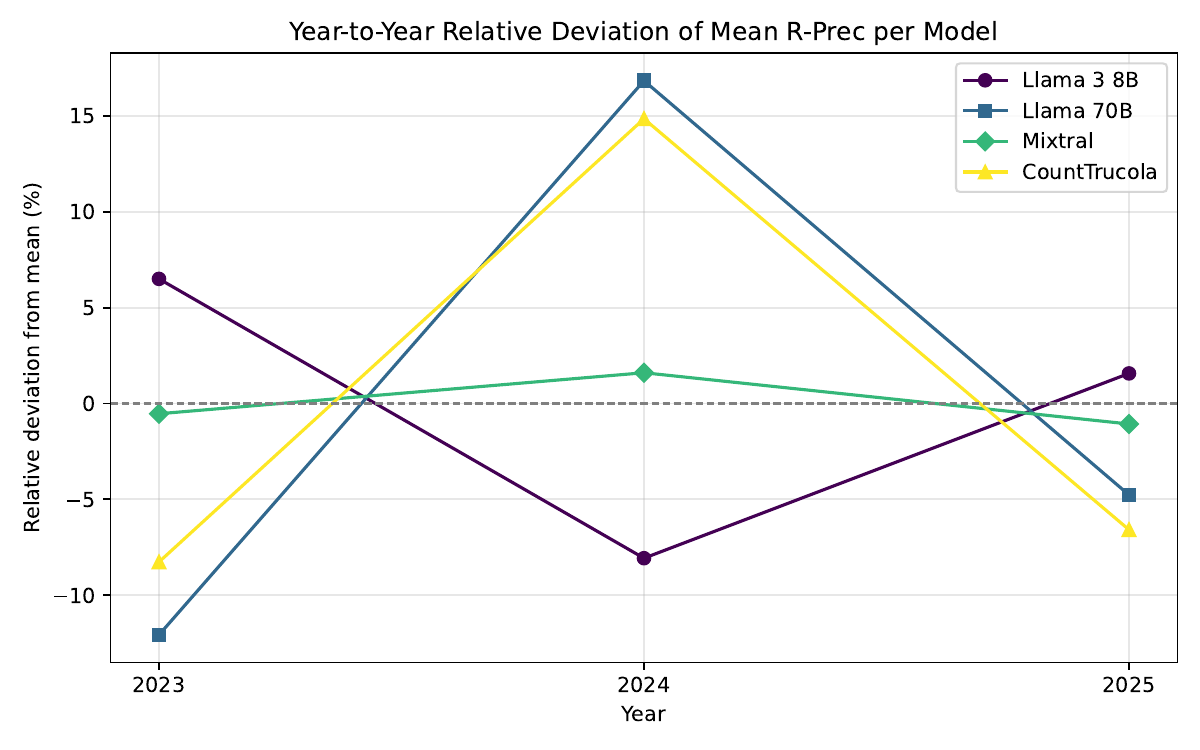}
 \end{subfigure}    
    \caption{Per-year performance on Mean R-Precision (top) and relative deviation from each model's three-year mean (bottom), evaluated on a subset of 50 artists  and 50 festivals appearing in all three years. }
    \label{fig:preexp}
\end{figure*}

Figure~\ref{fig:preexp} (top) reports Mean R-Precision per year and model. CountTRuCoLa consistently outperforms all LLM-based approaches across all years, 
consistent with the main results in Section~~\ref{sec:results}. To analyze year-to-year differences, we compute the relative deviation of each model's per-year score from its three-year 
mean, shown in Figure~\ref{fig:preexp} (bottom). The deviations are comparable in magnitude across LLMs and CountTRuCoLa alike: Llama 3.3 70B, Mixtral, and CountTRuCoLa 
peak in 2024, while Llama 3.1 8B reaches its lowest score in that year. Crucially, no model exhibits systematically stronger performance in 2023 or 2024 relative to 2025, 
and the LLMs do not show unusually large temporal fluctuations compared to CountTRuCoLa.

We find no evidence that the knowledge cutoff substantially degrades 2025 performance, and therefore proceed with 2025 data in all subsequent experiments.

\end{document}